\documentclass[letterpaper]{article} 
\usepackage{aaai2027}  
\usepackage[hyphens]{url}  
\usepackage{graphicx} 
\usepackage{natbib}  
\usepackage{caption} 
\usepackage{algorithm}
\usepackage{algorithmic}
\usepackage{booktabs}
\usepackage{multirow}
\usepackage{amsmath}
\usepackage{amsfonts}
\usepackage[table,xcdraw]{xcolor}
\definecolor{no1}{HTML}{FFB2B2}  
\definecolor{no2}{HTML}{FFD9B2}  

\usepackage{newfloat}
\usepackage{listings}
\DeclareCaptionStyle{ruled}{labelfont=normalfont,labelsep=colon,strut=off} 
\floatstyle{ruled}
\newfloat{listing}{tb}{lst}{}
\floatname{listing}{Listing}

\usepackage{booktabs}

\title{AdaGS: Confidence-Gated Diffusion Adaptation for Unposed Sparse-View 3D Gaussian Reconstruction}
\author{
    Beizhen Zhao\textsuperscript{\rm 1}\equalcontrib,
    Sicheng Yu\textsuperscript{\rm 1}\equalcontrib,
    Guanzhi Ding\textsuperscript{\rm 1},
    Yu Hu\textsuperscript{\rm 1}, 
    Hao Wang\textsuperscript{\rm 1}\corresponding
}

\affiliations{
    \textsuperscript{\rm 1}The Hong Kong University of Science and Technology (Guangzhou)\\

    bzhao610@connect.hkust-gz.edu.cn,
    syu605@connect.hkust-gz.edu.cn,
    haowang@hkust-gz.edu.cn
}

\begin{document}

\maketitle

\begin{abstract} 
Sparse-view outdoor 3D reconstruction remains challenging for Gaussian splatting.
Although diffusion restorers can provide additional pseudo-view supervision, we identify a failure pattern in diffusion-guided reconstruction: an unguided diffusion model may misinterpret floating artifacts and complete them into plausible but non-existent scene structures. 
To mitigate this challenge, we propose a confidence-gated diffusion adaptation framework for unposed sparse-view reconstruction. 
We introduce an artifact-aware adapter that is trained in two stages on our constructed Gaussian artifact dataset. 
The adapter suppresses floater-induced visual cues before diffusion restoration, while bidirectional confidence fusion accepts generated regions only when they are supported by cross-view geometric evidence. 
Furthermore, we introduce a Gaussian stabilization strategy to attenuate Gaussians that exhibit weak multi-view inconsistency, preventing them from repeatedly triggering hallucinations in later optimization. 
Extensive experiments on three outdoor datasets demonstrate that our method improves rendering fidelity and pose accuracy over existing approaches. 
\end{abstract}

\section{Introduction}

Reconstructing large-scale 3D scenes from unposed sparse images is a fundamental yet difficult problem in computer vision due to limited overlap, weak triangulation cues, and unreliable correspondence constraints~\cite{kerbl20233dgs,lin2025longsplat, yu2025opengs-slam, liu20243dgs-enhancer}. 
The difficulty becomes more severe in outdoor driving scenes, where the high speed motions pattern results in wide baselines, occlusions, textureless regions, and large depth variations and breaks the assumptions of conventional SfM pipelines~\cite{zhang2025transplat, zheng2025nexusgs}.
Recovering accurate camera poses and high-fidelity scenes from such limited observations therefore remains an important challenge~\cite{fu2024cf-3dgs,jiang2024cogs,shi2025trackgs}.

3D Gaussian Splatting (3DGS) has recently shown remarkable performance in novel view synthesis. 
However, its optimization is highly dependent on sufficient multi-view constraints and accurate camera initialization. 
Under unposed sparse-view settings, early Gaussian optimization is severely under-constrained~\cite{li2024dngaussian,xu2025depthsplat,zhang2025transplat}. 
The resulting intermediate representation often contains unstable depth distributions and view-dependent artifacts. 
Since they are projected into rendered views, they become part of the visual evidence used by subsequent reconstruction or restoration modules.

A natural strategy is to employ pretrained diffusion models to restore degraded renderings or synthesize pseudo-views~\cite{liu20243dgs-enhancer,wu2025difix3d+,yu2024viewcrafter}. 
However, directly applying diffusion restoration to outdoor unposed sparse 3DGS renderings introduces a subtle failure mode. 
A diffusion model is trained to explain visual patterns as meaningful image structures. 
When the input rendering contains floating Gaussian artifacts, the model may interpret them as valid object-level or scene-level cues, completing them into plausible but non-existent cars, trees, or building parts. 
Although such outputs may appear reasonable in a single view, they are not supported by the underlying multi-view consistency.

We observe a recurrent failure pattern in diffusion-guided sparse reconstruction.
During the optimization stage, unstable 3D Gaussians are first rendered as 2D artifacts. 
A diffusion model then transforms these artifacts into hallucinated but visually plausible content. 
When pseudo-views are directly used as supervision, the hallucinated regions are absorbed into the Gaussian representation and further corrupt the subsequent optimization. 
Based on this observation, the central question has changed from simply using diffusion models to extend sparse views to how to safely adapt a strong generative prior to an uncertain reconstruction process.

To address these challenges, we propose a confidence guided diffusion adaptation framework. 
First, we introduce a lightweight artifact-aware adapter.
The adapter is first trained on our constructed Gaussian artifact dataset, enabling it to distinguish sparse-reconstruction artifacts from valid scene structures. 
Then it is jointly optimized with the diffusion backbone. 
This design preserves the general restoration capability of the pretrained diffusion model and adapts only the task-specific conditioning interface needed for sparse 3D reconstruction.

In addition, we introduce bidirectional confidence fusion to prevent unsupported generated content from entering gaussian optimization. 
For each restored pseudo-view, we verify whether its pixels are supported by bidirectional neighboring reference frames through structure consistency. 
Only regions with reliable cross-view evidence are used as supervision. 
Finally, we propose a representation-side Gaussian stabilization strategy that identifies diffuse, weakly supported Gaussians and attenuates them during optimization.

Our main contributions are summarized as follows:
\begin{itemize}
    \item We develop a two-stage trained artifact-aware adapter that collaborates with a frozen diffusion backbone to enable reliable pseudo-view restoration.
    \item We introduce bidirectional confidence fusion to gate diffusion-restored pseudo-views according to cross-view geometric support, preventing unsupported hallucinations from corrupting joint optimization.
    \item We design a representation-side Gaussian stabilization strategy that suppresses diffuse artifact reservoirs in the 3D Gaussian field, further reducing the chances of triggering 2D hallucinations.
\end{itemize}

\section{Related Work}

\subsection{Unposed 3D Gaussian Splatting}
While 3DGS \cite{kerbl20233dgs} offers high-fidelity rendering, its reliance on COLMAP \cite{schonberger2016sfm} poses has motivated calibration-free alternatives.
Methods like CF-3DGS \cite{fu2024cf-3dgs}, TrackGS \cite{shi2025trackgs}, COGS \cite{jiang2024cogs}, and Rob-GS \cite{dong2025rob-gs} achieve unposed reconstruction by progressively refining cameras and Gaussians via dense 2D correspondences, pairwise registration, or joint optimization.
Currently, 3DGS-SLAM frameworks \cite{matsuki2024monogs,yu2025opengs-slam,sandstrom2025splat-slam,cheng2025s3po-gs} enable online unposed reconstruction from streaming images.
Recent works also leverage large pretrained models for initialization and alignment: InstantSplat \cite{fan2024instantsplat} uses DUSt3R \cite{wang2024dust3r}, whereas LongSplat \cite{lin2025longsplat} and \cite{cheng2025unposed} utilize MASt3R \cite{leroy2024mast3r} and VGGT \cite{wang2025vggt} with sub-graph or incremental optimization to efficiently scale to long sequences.

\subsection{Diffusion-guided 3Dgs from Sparse Views}
Sparse input inherently degrades novel view synthesis quality, prompting various regularization strategies \cite{kerbl20233dgs,lu2024scaffold-gs,zhao2025wavelet}.
Existing works mitigate floaters and artifacts by incorporating external geometric and depth priors \cite{li2024dngaussian,xu2025depthsplat,zhang2025transplat,paliwal2024coherentgs,zheng2025nexusgs,chung2024depth-regular}, generative models \cite{liu20243dgs-enhancer,wu2025difix3d+,zhong2025taming-video,yu2024viewcrafter,liu2024reconx}, photometric and opacity constraints \cite{han2024binocular,zhang2024cor-gs}, or stochastic density modulation \cite{xu2025dropoutgs,park2025dropgaussian,chen2025quantifying}.
However, these approaches typically rely on known COLMAP poses or remain object-centric.
While recent efforts tackle unposed or sparse-view settings, they face specific limitations: RegGS \cite{cheng2025reggs} utilizes NoPoSplat \cite{ye2024noposplat} but focuses on indoor scenes; Gscenes \cite{paul2024gaussianscene} employs diffusion and RGB-D priors but suffers from offline inefficiency; and SGD \cite{yu2025sgd} requires known poses and LiDAR for street scenes.
To overcome these limitations, we adapt the input interface while keeping the diffusion backbone frozen, (ii) admit only restoration residuals supported by geometry derived from real reference views, and (iii) suppress weakly supported Gaussians before they re-trigger restoration.

\begin{figure*}[t]
\centerline{\includegraphics[width=1.0\textwidth, trim=0cm 8.3cm 10.7cm 0cm, clip]{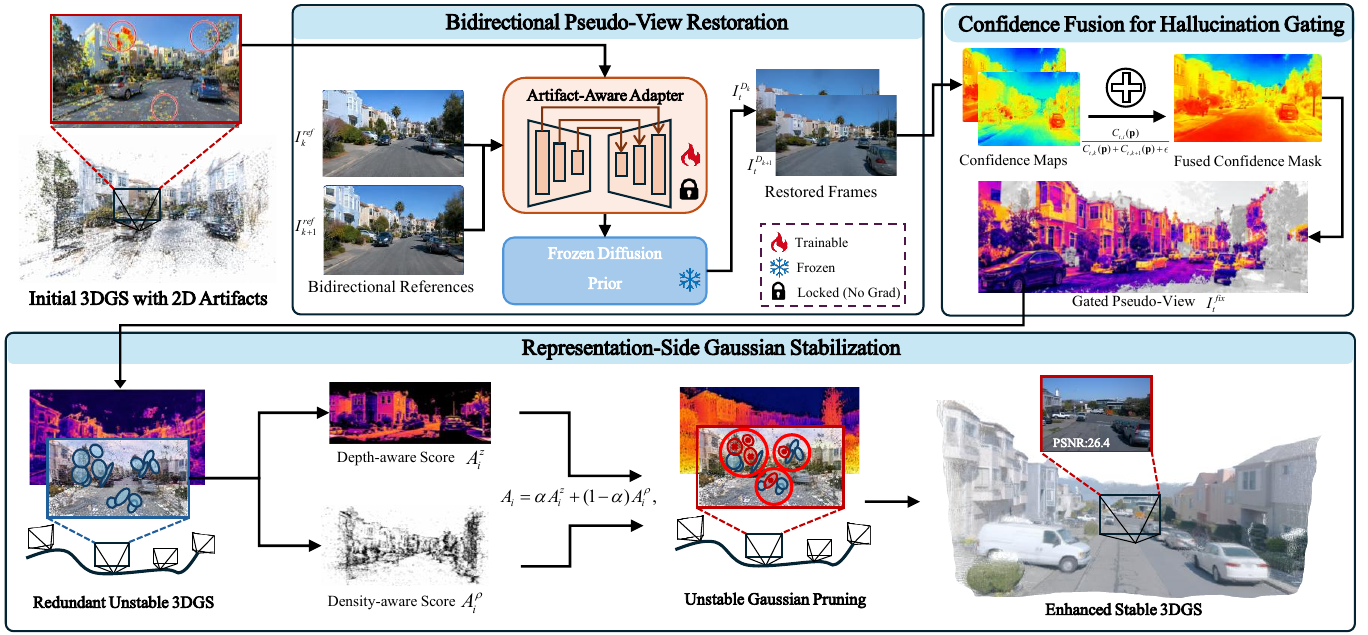}}
\caption{\textbf{Overview of the proposed framework.} We begin by the sparse input views and through the cross-view vision to select and restore frame.
We design a lightweight artifact-aware adapter to denoise the Gaussian rendered image and utilize a diffusion-based model for image completion.
After that, we combine the two fixed images by calculating reprojection overlap score and conduct feature mapping to generate a confidence mask to guide the joint optimization.}
\label{framework}
\end{figure*}

\section{Methodology}

\subsection{Problem Formulation and Artifact Feedback}
\label{sec:problem}

Given $N$ sparse RGB images with unknown poses $\mathcal{T}$, we represent the scene using a set of 3D Gaussians $\mathcal{G}=\{(\bf{\mu}_j, \bf{\Sigma}_j, \bf{c}_j, o_j)\}$. Poses and real-view depth priors are initialized via MASt3R. At optimization round $r$, we render an RGB-D pair at a sampled target pose $T_t^{(r)}$:
\begin{equation}
I_{t}^{\mathrm{gs},(r)}, D_{t}^{\mathrm{gs},(r)} = \mathcal{R}\left(\mathcal{G}^{(r)},T_t^{(r)}\right).
\label{eq:gs_render}
\end{equation}

A diffusion-based restoration module then generates a pseudo-target $\widetilde I_t^{(r)} = \mathcal{D}_{\phi}\left(I_t^{\mathrm{gs},(r)}, I^{\mathrm{ref}}\right)$ to supervise the joint update of $\mathcal{G}$ and $\mathcal{T}$. 
However, ungated diffusion often converts rendered 3D floaters into visually plausible but geometrically erroneous content. Using this pseudo-target as supervision absorbs these errors back into $\mathcal{G}$, creating an artifact loop. 
We mitigate this loop via three complementary mechanisms: artifact-aware input adaptation, geometry-gated pseudo-view restoration, and representation-side Gaussian stabilization.

\subsection{Two-Stage Artifact-Aware Adapter Training}

We introduce a lightweight artifact-aware adapter $A_{\theta}$ to suppress floater-induced cues before diffusion restoration. 
The adapter receives the current Gaussian-rendered image and two neighboring reference images:
\begin{equation}
    \hat{I}_t
    =
    A_{\theta}
    \left(
    I_{k}^{ref},
    I_t^{gs},
    I_{k+1}^{ref}
    \right),
\end{equation}
where $\hat{I}_t$ is an artifact-suppressed intermediate image. 
The adapter is implemented as a lightweight encoder-decoder network with skip connections. 
Instead of treating it as a generic image deblurring network, we use it as a reconstruction-specific conditioning interface between uncertain 3DGS renderings and a frozen diffusion restoration prior.

\paragraph{Stage I: Direct adapter pretraining.}
We first train $A_{\theta}$ on a constructed Gaussian artifact dataset:
\begin{equation}
    \mathcal{D}_{ada}
    =
    \left\{
    \left(
    I^{gs},
    I^{-}_{ref},
    I^{+}_{ref},
    I^{*}
    \right)
    \right\},
\end{equation}
where $I^{gs}$ is a corrupted Gaussian-rendered image, $I^{-}_{ref}$ and $I^{+}_{ref}$ are neighboring reference images, and $I^{*}$ denotes the corresponding clean target. This dataset exposes the adapter to typical sparse-reconstruction artifacts, including floating Gaussians, ghosting, incomplete structures, blurred boundaries, and depth-induced distortions.

The training objective combines pixel-level reconstruction and structural similarity losses, encouraging the adapter to remove unreliable floater cues while preserving scene structures supported by the neighboring references.

\paragraph{Stage II: Diffusion-coupled adapter tuning.}
After direct pretraining, we cascade the adapter with the pretrained diffusion restoration model $\mathcal{D}_{\phi}$. In this stage, the diffusion backbone is frozen and only the adapter parameters $\theta$ are updated.

Given the adapter output $\hat{I}_t$, the frozen diffusion model produces restoration candidates conditioned on neighboring references:
\begin{equation}
    I_t^{D_k}
    =
    \mathcal{D}_{\phi}^{frozen}
    \left(
    \hat{I}_t,
    I_k^{ref}
    \right),
    \quad
    I_t^{D_{k+1}}
    =
    \mathcal{D}_{\phi}^{frozen}
    \left(
    \hat{I}_t,
    I_{k+1}^{ref}
    \right).
\end{equation}
This stage teaches the adapter to produce an intermediate image that is not merely clean in pixel space, but also compatible with the frozen diffusion restoration prior. 
In other words, the adapter learns how to present sparse 3DGS renderings to the diffusion model in a safer form, reducing the chance that floating artifacts are interpreted as valid object-level cues.
After Stage II, both $A_{\theta}$ and $\mathcal{D}_{\phi}$ are frozen and reused as a fixed restoration module during scene reconstruction.

\begin{figure*}[t]
\centerline{\includegraphics[width=1.0\textwidth, trim=0cm 14.2cm 9.8cm 0cm, clip]{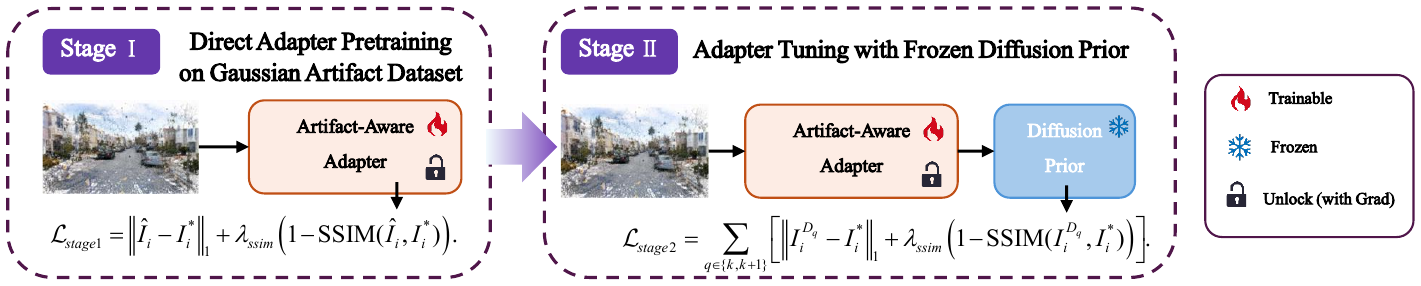}}
\caption{\textbf{Two-Stage Artifact-Aware Adapter Training.} The adapter is first trained on our Gaussian Artifact Dataset. Then in stage 2, the adapter is trained with a frozen diffusion to adapt the diffusion prior.
}
\label{framework}
\end{figure*}

\subsection{Bidirectional Adapter-Diffusion Restoration}

During scene reconstruction, we render the current Gaussian field into a target view and restore it using the adapter-diffusion module. 
Given $I_t^{gs}$ and two selected reference images $I_k^{ref}$ and $I_{k+1}^{ref}$, we first compute:
\begin{equation}
    \hat{I}_t
    =
    A_{\theta}
    \left(
    I_k^{ref},
    I_t^{gs},
    I_{k+1}^{ref}
    \right).
\end{equation}
Then, the frozen diffusion model generates two reference-conditioned restoration candidates:
\begin{equation}
    I_t^{D_k}
    =
    \mathcal{D}_{\phi}
    \left(
    \hat{I}_t,
    I_k^{ref}
    \right),
    \quad
    I_t^{D_{k+1}}
    =
    \mathcal{D}_{\phi}
    \left(
    \hat{I}_t,
    I_{k+1}^{ref}
    \right).
\end{equation}

The adapter removes artifact cues that are likely to trigger hallucination, while the diffusion model contributes its pretrained restoration prior without updating its parameters.
The bidirectional design reduces the bias caused by relying on a single reference view. 
Regions that are ambiguous in one direction may still be supported by the other direction, while hallucinated details that appear only under one reference can be identified and suppressed in the following confidence fusion step.

\subsection{Reference Selection and Overlap Estimation}

For each target pseudo-view, we select two reference frames with sufficient cross-view consistency. 
The selection is based on visual overlap rather than temporal adjacency, making the method applicable to unordered sparse image collections. 
In practice, we estimate overlap by projecting the current rendered depth from one view to another and evaluating whether the projected pixels are visible and depth-consistent.

We summarize the overlap reliability between two views $a$ and $b$ as:
\begin{equation}
    C_{ab}(\mathbf{p})
    =
    \mathcal{O}_{ab}(\mathbf{p})
    \cdot
    s_d(\mathbf{p})
    \cdot
    s_t,
\end{equation}
where $\mathcal{O}_{ab}$ denotes the valid visibility overlap, $s_d$ measures depth consistency after reprojection, and $s_t$ down-weights references with excessive viewpoint displacement.
Detailed computational process can be found in the Supplementary Material. 
A higher $C_{ab}$ indicates that the corresponding reference view provides stronger geometric support for the target pseudo-view. 
This overlap confidence is later used to fuse bidirectional diffusion outputs.

\subsection{Confidence Fusion for Hallucination Gating}

The two restored candidates $I_t^{D_k}$ and $I_t^{D_{k+1}}$ may contain different details because they are conditioned on different reference frames. We therefore fuse them through residual confidence fusion. Let
\begin{equation}
    R_t^{k}
    =
    I_t^{D_k}
    -
    I_t^{gs},
    \quad
    R_t^{k+1}
    =
    I_t^{D_{k+1}}
    -
    I_t^{gs}
\end{equation}
be the restoration residuals. The fused pseudo-view is
\begin{equation}
    I_t^{fix}
    =
    I_t^{gs}
    +
    W_k \odot R_t^{k}
    +
    W_{k+1} \odot R_t^{k+1},
\end{equation}
where
\begin{equation}
    W_i(\mathbf{p})
    =
    \frac{
    C_{t,i}(\mathbf{p})
    }
    {
    C_{t,k}(\mathbf{p})
    +
    C_{t,k+1}(\mathbf{p})
    +
    \epsilon
    },
    \quad
    i\in\{k,k+1\},
\end{equation}
where we infer the confidence mask $C_m$ using bidirectional consistency verification. 
We compute correspondence sets $\mathcal{M}_{k}$ and $\mathcal{M}_{k+1}$ between the fused pseudo-view and the two neighboring reference frames. 
A pixel is considered reliable only when its generated content is supported by cross-view correspondences:
\begin{equation}
C_m(\mathbf{p}) = \frac{1}{2} \left( \mathbb{I}_{\mathcal{M}_{k}}(\mathbf{p}) + \mathbb{I}_{\mathcal{M}_{k+1}}(\mathbf{p}) \right)
\end{equation}
This mask acts as a hallucination gate: generated regions are allowed to supervise reconstruction only when they are supported by cross-view consistency.

\subsection{Representation-Side Gaussian Stabilization}

While confidence fusion suppresses pixel-level hallucinations, existing 3D floaters can still be re-rendered and repeatedly trigger diffusion errors. 
We therefore introduce a Gaussian stabilization strategy. 
Intuitively, valid surfaces form dense, depth-stable structures, whereas floaters are sparse and depth-inconsistent.
Based on this observation, we compute an artifactness score to identify and attenuate problematic Gaussians.

For the depth-aware term, we partition Gaussians into depth clusters using quantile boundaries. 
Let $z_i$ be the depth of Gaussian $g_i$ under the current visible views. 
We compute quantile boundaries from the empirical depth distribution.
The cluster assignment of $g_i$ is denoted by $c(i)$. 
Within each cluster, Gaussians with inconsistent or weak depth support receive higher depth artifactness:
\begin{equation}
    A_i^{z}
    =
    1-
    \operatorname{Norm}
    \left(
    \frac{
    \sigma_z(i)
    }
    {
    \bar{v}_i+\epsilon
    }
    \right),
\end{equation}
where $\sigma_z(i)$ measures the depth variation of $g_i$ across visible views, and $\bar{v}_i$ denotes its normalized visibility frequency.

For the density-aware term, we estimate the local density around each Gaussian and compute a normalized density-dispersion score:
\begin{equation}
    A_i^{\rho}
    =
    1-
    \operatorname{Norm}(\sum_{g_j\in\mathcal{N}(g_i)}
    \exp
    \left(
    -
    \frac{
    \left\|
    \boldsymbol{\mu}_i-\boldsymbol{\mu}_j
    \right\|_2^2
    }
    {2\sigma_{\rho}^{2}}
    \right)),
\end{equation}
where $\mathcal{N}(g_i)$ denotes the local neighborhood of $g_i$. 
Gaussians in sparse and diffuse regions receive higher density artifactness, reflecting the weak spatial support of floating artifacts.

During training, we attenuate the opacity of Gaussians with higher score softly: 
\begin{equation}
    A_i
    =
    \alpha A_i^{z}
    +
    (1-\alpha)A_i^{\rho},
\end{equation}
Gaussians that are visible inconsistently across views or located in sparse, diffuse regions receive higher artifactness scores and will be pruned stochastically for stabilization.
This strategy reduces the 3D floaters that would be repeatedly rendered and triggered by the diffusion model.

\begin{table*}[t]
\caption{\textbf{Novel View Synthesis Result on Three Dataset.} Methods marked with $^*$ fail on some scenes under the extremely sparse-view setting; results are reported only over successfully reconstructed scenes.}
\vspace{-2pt}
\centering
{
\small
\setlength{\tabcolsep}{6pt} 
\renewcommand{\arraystretch}{1}  
\begin{tabular}{l|ccc|ccc|ccc}
\toprule
\multirow{2}{*}{\textbf{Method}} & \multicolumn{3}{c|}{DL3DV \cite{ling2024dl3dv}}  & \multicolumn{3}{c|}{Waymo \cite{Sun_2020_CVPRwaymo}} & \multicolumn{3}{c}{KITTI \cite{geiger2013kittidata}} \\ \cmidrule(lr){2-10}
\multicolumn{1}{c|}{} & PSNR↑ & SSIM↑ & \multicolumn{1}{c|}{LPIPS↓} & PSNR↑ & SSIM↑ & \multicolumn{1}{c|}{LPIPS↓} & PSNR↑ & SSIM↑ & LPIPS↓ \\ \midrule
CF-3DGS \cite{fu2024cf-3dgs} & 13.01 & 0.426 & 0.582 & 13.67 & 0.590 & 0.554 & 9.79 & 0.303 & 0.619 \\
$\text{VideoLifter}^*$ \cite{cong2025videolifter} & 17.84 & 0.519 & 0.433 & 13.53 & 0.471 & 0.592 & 8.356 & 0.194 & 0.647 \\
InstantSplat \cite{fan2024instantsplat} & 10.88 & 0.351 & 0.615 & 12.49 & 0.564 & 0.645 & 9.926 & 0.367& 0.654\\
LongSplat \cite{lin2025longsplat}& 17.40 & 0.516 & 0.428 & 15.06 & 0.551 & 0.522 & 10.69 & 0.285 &0.617 \\
$\text{RegGS}^*$ \cite{cheng2025reggs} & 19.77 & 0.612 & 0.391 & - & - & - & - & - & -  \\
$\text{OpenGS-SLAM}$ \cite{yu2025opengs-slam} & 16.08 & 0.428 & 0.403 & 21.37 & 0.765& 0.457 & 13.87 & 0.433 & 0.537 \\
$\text{AnySplat}$ \cite{jiang2025anysplat} & \cellcolor{no2}21.26 & \cellcolor{no2}0.664 & 0.309 & 17.40 & 0.592& 0.461 & 11.60 & 0.371 & 0.576 \\
$\text{ProSplat}$ \cite{lu2025prosplat} & 20.37 & 0.642 & 0.266 & 17.59 & 0.582& 0.479 & 11.24 & 0.346	& 0.592 \\
$\text{Depth Anything3}$ \cite{lin2025depth} & 17.64 & 0.519 & \cellcolor{no2}0.196 & 19.39& 0.731& \cellcolor{no1}0.317 & 12.39 & 0.383 & 0.538 \\
$\text{DGGT}$ \cite{chen2025dggt} & 16.65 & 0.535 & 0.379 & 20.91 & 0.634& 0.350 & 14.89 & 0.397 & 0.546 \\
$\text{S3PO-GS}$ \cite{cheng2025s3po-gs} & 20.67 & 0.647 & 0.219 & \cellcolor{no2}22.03 & \cellcolor{no2}0.771 & 0.452 & \cellcolor{no2}15.58 & \cellcolor{no2}0.510 & \cellcolor{no2}0.513 \\
\midrule
Ours & \cellcolor{no1}\textbf{24.27} & \cellcolor{no1}\textbf{0.753} & \cellcolor{no1}\textbf{0.183} &  \cellcolor{no1}\textbf{23.76} & \cellcolor{no1}\textbf{0.777} & \cellcolor{no2}\textbf{0.348} & \cellcolor{no1}\textbf{17.95} & \cellcolor{no1}\textbf{0.605} & \cellcolor{no1}\textbf{0.472}\\ \bottomrule
\end{tabular}
}

\label{tab:main}
\vspace{-5pt}
\end{table*}

\begin{figure*}[t]
\centerline{\includegraphics[width=1\textwidth]{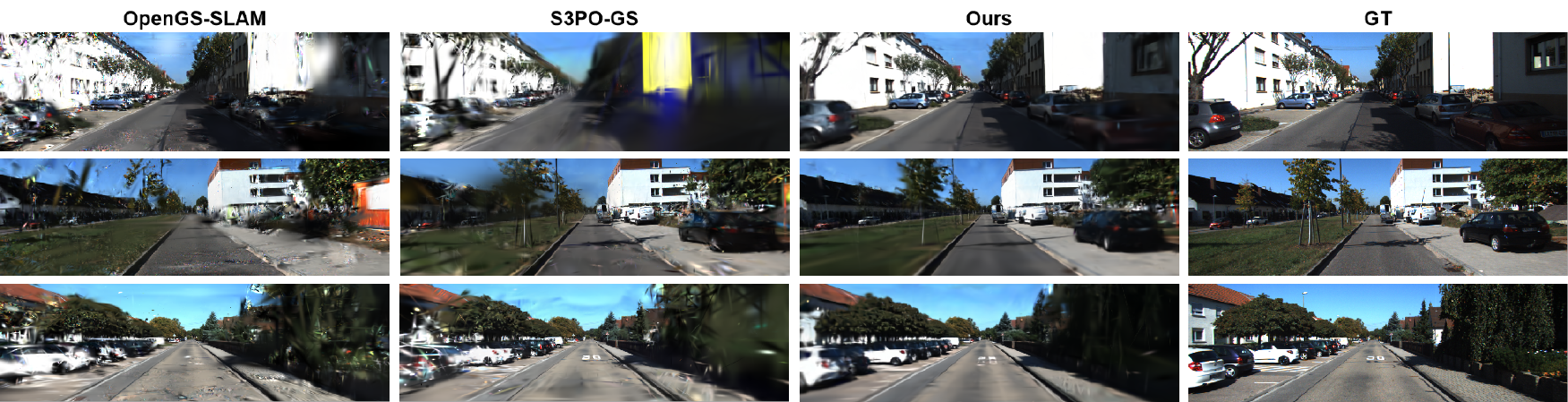}}
\vspace{-2pt}
\caption{\textbf{Visual quality on KITTI dataset.} Our method produces cleaner and more view-consistent renderings under challenging outdoor conditions, including large viewpoint variations, complex illumination, and distant scene content. 
Competing methods exhibit more visible artifacts, blurring, and inconsistent appearance, particularly in distant or poorly constrained regions.}
\vspace{-5pt}
\label{fig:kitti1}
\end{figure*}

\subsection{Joint Pose and Gaussian Optimization}

During scene reconstruction, the adapter $A_{\theta}$ and the diffusion model $\mathcal{D}_{\phi}$ are fixed. 
We optimize only scene-specific variables, including camera poses, Gaussian parameters, and exposure correction. 
The optimization uses both real input views and confidence-gated pseudo-views.

Camera poses are refined through Lie algebra updates:
\begin{equation}
    T_t
    \leftarrow
    \exp(\Delta \boldsymbol{\xi}_t)T_t,
\end{equation}
where $\Delta \boldsymbol{\xi}_t\in\mathfrak{se}(3)$ is the pose update. To handle photometric variation, we use a per-view affine exposure correction:
\begin{equation}
    I_t^{corr}
    =
    a_t I_t + b_t.
\end{equation}

The overall optimization objective is: 
\begin{equation}
    \mathcal{L}
    =
    \mathcal{L}_{real}
    +
    \lambda_{pv}\mathcal{L}_{pv}
    +
    \lambda_{d}\mathcal{L}_{d}
    +
    \lambda_{s}\mathcal{L}_{scale}.
\end{equation}
Here, $\mathcal{L}_{real}$ is the photometric supervision from the original sparse input views. 
$\mathcal{L}_{pv}$ uses the fused pseudo-view $I_t^{fix}$ as additional supervision, but is weighted by the confidence mask $C_m$ so that unsupported generated  regions do not contribute to optimization. 
$\mathcal{L}_{d}$ provides depth consistency when reliable pseudo-depth is available, and $\mathcal{L}_{scale}$ regularizes Gaussian scale to prevent excessive expansion.

\section{Experiments}
\label{Experiments}

\subsection{Experiments Setup}

\paragraph{Datasets.}
We evaluate our method on three widely used outdoor reconstruction datasets with increasing levels of viewpoint and appearance variation:
\begin{itemize}
  \item \textbf{DL3DV \cite{ling2024dl3dv} (easy):} an outdoor benchmark with relatively dense multi-view coverage and moderate viewpoint changes.
  \item \textbf{Waymo \cite{Sun_2020_CVPRwaymo} (moderate):} a large-scale driving dataset containing urban and highway scenes with varying illumination and substantial inter-frame motion.
  \item \textbf{KITTI \cite{geiger2013kittidata} (hard):} a driving dataset characterized by large viewpoint changes, strong motion parallax, and extensive low-texture regions.
\end{itemize}

The original sequences are densely sampled, with an average adjacent-view overlap above $0.98$. 
To construct a challenging sparse-view setting, we uniformly downsample each sequence until the average overlap decreases to approximately $0.75$. 
For KITTI and Waymo, we use the first 200 frames of each sequence and sample every tenth frame, resulting in 20 input views; the remaining 180 frames are used for novel-view evaluation. 
For DL3DV, we sample every 30th frame from a 300-frame sequence, yielding 10 input views.

\paragraph{Metrics.}
We evaluate novel view synthesis using PSNR, SSIM, and LPIPS, and assess pose accuracy using the absolute trajectory error root mean square error (ATE RMSE). 
Additional detailed results and GPU memory consumption are provided in the supplementary material.

\paragraph{Baselines.}
We compare our method with representative state-of-the-art approaches for unposed sparse-view reconstruction. 
These include optimization-based methods: CF-3DGS \cite{fu2024cf-3dgs}, RegGS \cite{cheng2025reggs}, VideoLifter \cite{cong2025videolifter}, InstantSplat \cite{fan2024instantsplat}, LongSplat \cite{lin2025longsplat}, OpenGS-SLAM \cite{yu2025opengs-slam}, and S3PO-GS \cite{cheng2025s3po-gs}, as well as feed-forward methods: AnySplat \cite{jiang2025anysplat}, ProSplat \cite{lu2025prosplat}, Depth Anything 3 \cite{lin2025depth}, and DGGT \cite{chen2025dggt}.

\paragraph{Implementation Details.}
Our reconstruction experiments are implemented in PyTorch and run on a single NVIDIA A6000 GPU. 
We use the publicly available pretrained Difix3D \cite{wu2025difix3d+} model as the diffusion restoration backbone and MASt3R \cite{leroy2024mast3r} for pose initialization.

The artifact-aware adapter is trained on over 10,000 samples, each containing a Gaussian-rendered artifact image, two neighboring reference images, and the corresponding clean target. 
Training is performed on four NVIDIA A6000 GPUs using AdamW and a cosine learning-rate schedule. 
The adapter is trained for 1,000 epochs and converges in approximately 10 hours.

\begin{table}[t!]
    \vspace{-2pt}
    \caption{\textbf{Pose estimation results--ATE RMSE on Three Dataset.} 
    Methods marked with $^*$ fail on some scenes under the extremely sparse-view setting.
    }
    \centering
    \setlength{\tabcolsep}{2pt} 
    \renewcommand{\arraystretch}{1}  
    \begin{tabular}{l|c|c|c}
    \toprule
    \textbf{Method}  & DL3DV & Waymo & KITTI \\
    \midrule
    CF-3DGS~\cite{fu2024cf-3dgs} & 3.093 & 10.842 & 65.891 \\
    $\text{VideoLifter}^*$ \cite{cong2025videolifter} & 3.184 & 32.712 & 81.221 \\
    Instantsplat \cite{fan2024instantsplat} & 0.333 & 9.992 & 65.103 \\
    Longsplat \cite{lin2025longsplat} & 0.919 & 7.548 & 67.732 \\
    $\text{RegGS}^*$ \cite{cheng2025reggs}  & 2.045 & - & - \\
    $\text{OpenGS-SLAM}$ \cite{yu2025opengs-slam} & 1.586 & 1.784 & 4.722 \\
    $\text{Anysplat}$ \cite{jiang2025anysplat} & 0.413 & 1.908 & 6.213 \\
    $\text{Prosplat}$ \cite{lu2025prosplat} & 0.173 & 1.894 & 6.613 \\
    $\text{Depthanything3}$ \cite{lin2025depth} & 0.101 & 1.464 & 3.531 \\
    $\text{DGGT}$ \cite{chen2025dggt} & 0.243 & 1.864 & 4.102 \\
    $\text{S3PO-GS}$ \cite{cheng2025s3po-gs} & 0.343 & 3.149 & 4.490 \\
    \midrule
    \textbf{Ours} & \textbf{0.077} & \textbf{1.352} & \textbf{2.471} \\
    \bottomrule
    \end{tabular}
    \vspace{-5pt}
    \label{tab:ate}
\end{table}

\begin{figure}[t!]
  \centering
  \includegraphics[width=1.0\columnwidth]{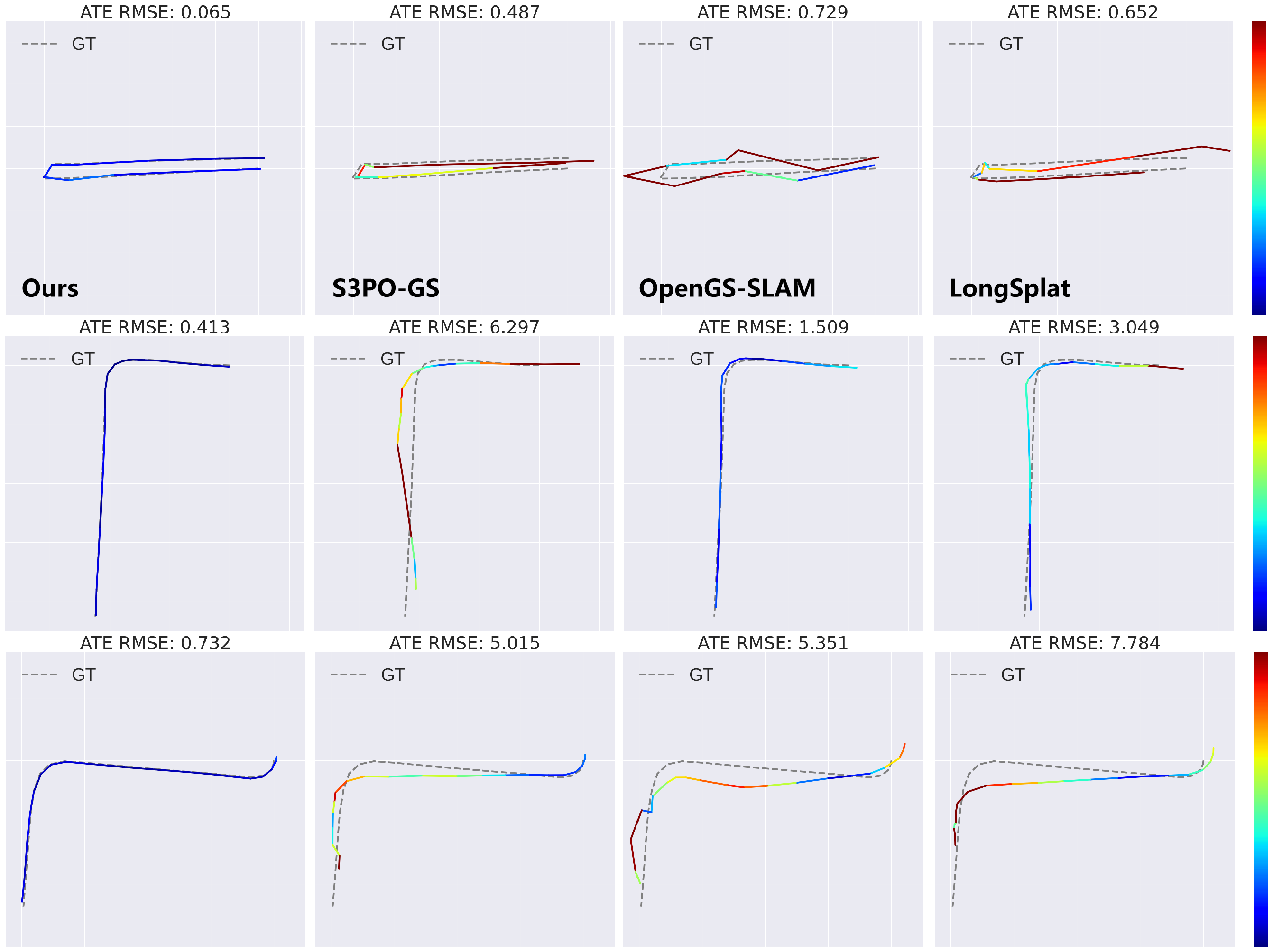}
  \vspace{-5pt}
  \caption{\textbf{Camera trajectory comparison on KITTI and DL3DV.}
    Our method achieves more accurate and robust camera pose estimation under sparse-view conditions, with reduced trajectory drift compared with competing methods.}
  \label{fig:camera-traj}
  \vspace{-5pt}
\end{figure}

\subsection{Experimental Results and Analysis}

Tables~\ref{tab:main} and~\ref{tab:ate} report the novel-view synthesis and camera pose estimation results, respectively. 
Qualitative rendering comparisons are shown in Fig.~\ref{fig:kitti1}, while Fig.~\ref{fig:camera-traj} visualizes the estimated camera trajectories.

\paragraph{Novel-view synthesis.}
Our method achieves the best PSNR and SSIM across three datasets, together with the best LPIPS on DL3DV and KITTI and the second-best LPIPS on Waymo. These results demonstrate improvements in rendering fidelity and perceptual quality under sparse, unposed observations.

On DL3DV, where the views retain relatively strong visual overlap, our method improves PSNR from 21.26 to 24.27. The qualitative comparisons show cleaner and more view-consistent renderings in boundary and distant regions.

Waymo contains substantial inter-frame motion, illumination variation, and partial occlusions. Under these conditions, direct diffusion restoration may introduce details that are plausible in a single view but inconsistent across neighboring observations. The artifact-aware adapter reduces misleading floater cues, while confidence fusion prevents unsupported generated regions from entering the optimization. Consequently, our method achieves the highest PSNR and SSIM on Waymo while remaining competitive in LPIPS.

KITTI is the most challenging setting because of large viewpoint changes, strong motion parallax, and extensive low-texture regions. Our method improves PSNR from 15.58 to 17.95 and SSIM from 0.510 to 0.605 over the strongest competing results. As shown in Fig.~\ref{fig:kitti1}, it produces cleaner and more consistent renderings in distant and poorly constrained regions, where competing methods exhibit visible artifacts and local distortions.

\begin{figure*}[t!]
\centerline{\includegraphics[width=1.0\textwidth]{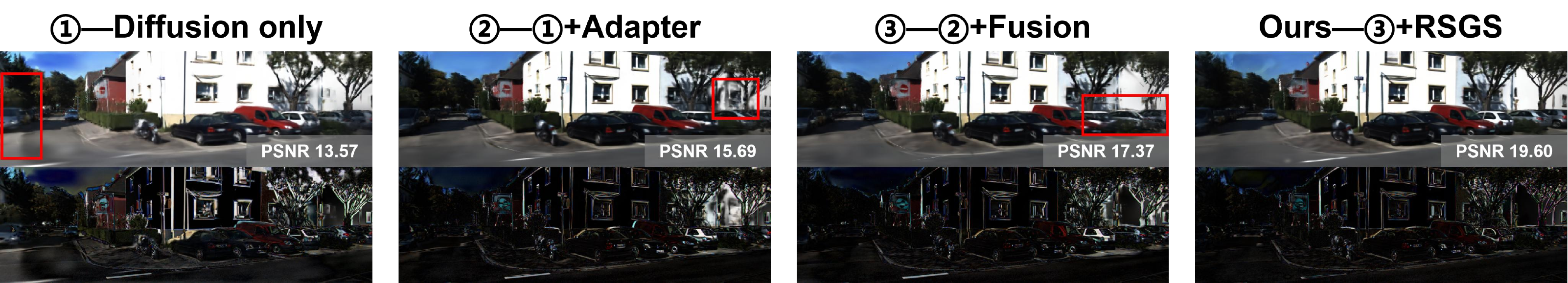}}
  \vspace{-5pt}
  \caption{\textbf{Qualitative component ablation.}
    Starting from direct diffusion supervision, the artifact-aware adapter reduces floater-induced artifacts, confidence fusion suppresses unsupported generated regions, and representation-side Gaussian stabilization (RSGS) further removes artifacts from the Gaussian field. The full model produces the cleanest and most consistent rendering.}
  \label{fig:ablation-gradual}
  \vspace{-5pt}
\end{figure*}

\paragraph{Pose estimation.}
As shown in Tab.~\ref{tab:ate}, our method achieves the lowest ATE RMSE on all three datasets. The improvement is particularly evident on Waymo and KITTI, where unreliable pseudo-view supervision may introduce incorrect photometric gradients and accumulated pose drift. By gating unsupported generated regions and attenuating unstable Gaussians, our method provides more reliable constraints for joint pose and Gaussian optimization. Figure~\ref{fig:camera-traj} further demonstrates more accurate and robust trajectory estimation with reduced drift under challenging sparse-view conditions.

\begin{table}[t!]
\caption{\textbf{Ablation Study on Waymo and KITTI.} Each component consistently improves novel-view synthesis quality, with the full model achieving the best overall performance.}
\vspace{-5pt}
\centering
{
\fontsize{9.5}{9.5}\selectfont  
\setlength{\tabcolsep}{1pt} 
\renewcommand{\arraystretch}{1}  
\begin{tabular}{l|ccc|ccc}
\toprule
\multirow{2}{*}{\textbf{Method}} & \multicolumn{3}{c|}{Waymo}  & \multicolumn{3}{c}{KITTI} \\ \cmidrule(lr){2-7}
\multicolumn{1}{c|}{} & PSNR↑ & SSIM↑ & \multicolumn{1}{c|}{LPIPS↓} & PSNR↑ & SSIM↑ & LPIPS↓ \\ \midrule
Diffusion only & 18.16 & 0.705 & 0.420 & 13.14 & 0.497 & 0.537 \\
w/o Adapter & 22.65 & 0.736 & 0.391 & 15.95 & 0.566 & 0.503 \\
w/o Fusion & 22.98 & 0.749 & 0.378 & 16.90 & 0.593 & 0.499 \\
w/o RSGS & 23.01 & 0.752 & 0.370 & 17.10 & 0.598 & 0.482 \\
\midrule
\textbf{Ours} & \textbf{23.76} & \textbf{0.777}& \textbf{0.348}&  \textbf{17.95} & \textbf{0.605} & \textbf{0.472} \\ \bottomrule
\end{tabular}
}
\vspace{-5pt}
\label{tab:ablation study}
\end{table}

\begin{figure}[t!]
  \centering
  \includegraphics[width=1\columnwidth]{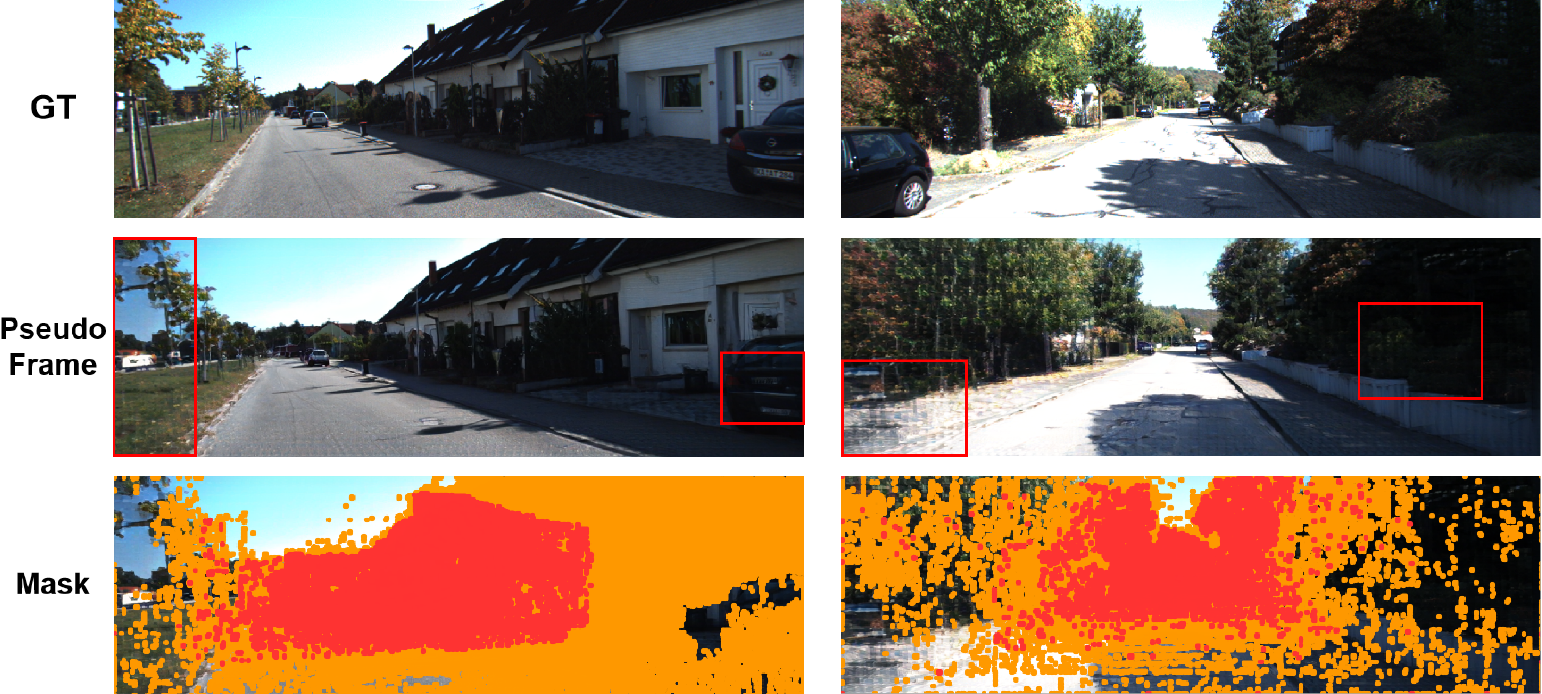}
  \vspace{-5pt}
  \caption{\textbf{Visualization of confidence mask.}
    The confidence mask identifies pseudo-view regions that lack reliable cross-view support. Excluding these low-confidence regions prevents unsupported diffusion content from entering the reconstruction objective.}
  \label{fig:ablation-confidence-mask}
  \vspace{-10pt}
\end{figure}

\subsection{Ablation Study}

\begin{figure}[t!]
  \centering
  \includegraphics[width=1\columnwidth]{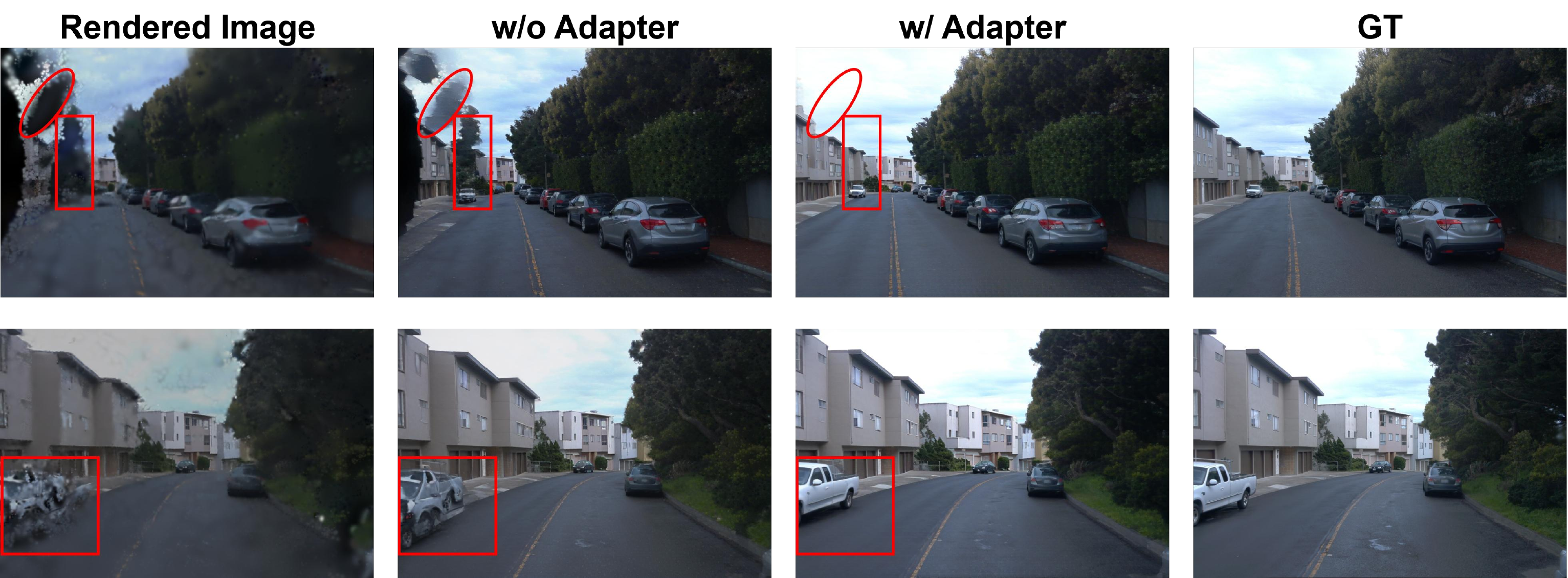}
  \vspace{-5pt}
  \caption{\textbf{Effect of the artifact-aware adapter.}
    The adapter suppresses floater-induced visual cues in Gaussian-rendered inputs before diffusion restoration, reducing hallucinated content and producing more reliable pseudo-views.
    }
  \label{fig:ablation-artifact-aware adapter}
  \vspace{-10pt}
\end{figure}

We conduct ablations on Waymo and KITTI to evaluate each component. In Tab.~\ref{tab:ablation study}, \textit{only diffusion} directly uses diffusion-restored pseudo-views without the proposed safeguards, while \textit{w/o Adapter}, \textit{w/o Fusion}, and \textit{w/o RSGS} remove the artifact-aware adapter, confidence fusion, and representation-side Gaussian stabilization, respectively. All variants use identical inputs, initialization, optimization schedules, and hyperparameters. Quantitative results are reported in Tab.~\ref{tab:ablation study}, with qualitative comparisons shown in Figs.~\ref{fig:ablation-gradual}, \ref{fig:ablation-confidence-mask}, and~\ref{fig:ablation-artifact-aware adapter}.

\paragraph{Diffusion only.}
Directly using diffusion-restored pseudo-views yields the weakest performance on both datasets. Although the generated images may appear plausible individually, unsupported content can be inconsistent across views and subsequently absorbed into the Gaussian representation. This confirms that direct diffusion supervision alone is insufficient for reliable sparse-view reconstruction.

\paragraph{Artifact-aware adapter.}
Removing the adapter causes a clear performance drop across all metrics. As shown in Fig.~\ref{fig:ablation-artifact-aware adapter}, the adapter suppresses floater-induced visual cues before diffusion restoration, reducing the likelihood that rendering artifacts are completed into plausible but incorrect content.

\paragraph{Confidence fusion.}
Without confidence fusion, restored regions contribute to pseudo-view supervision regardless of their cross-view support, allowing generated content in occluded or poorly constrained regions to enter the optimization. The confidence mask in Fig.~\ref{fig:ablation-confidence-mask} excludes such regions, improving PSNR from 22.98 to 23.76 on Waymo and from 16.90 to 17.95 on KITTI.

\paragraph{Representation-side Gaussian stabilization.}
Disabling Gaussian stabilization also degrades performance. Although confidence fusion filters unsupported pseudo-view regions, unstable floaters already stored in the Gaussian field can still be re-rendered and repeatedly trigger diffusion errors. The proposed depth- and density-aware artifactness score attenuates these persistent artifact sources, complementing image-space gating.

Overall, the three components interrupt the artifact amplification loop by suppressing artifact cues, rejecting unsupported content, and attenuating persistent 3D sources.

\subsection{Limitations}
Our method has several directions for further improvement. 
First, when view overlap is extremely limited or large regions lack texture, cross-view evidence may be insufficient to verify all restored regions. More robust correspondence and uncertainty estimation could improve confidence fusion in these cases. 
Second, while the artifact-aware adapter is trained on diverse Gaussian artifacts, broader artifact simulation could further improve its generalization to unseen reconstruction errors. 
Third, our current formulation assumes static scenes; extending it to dynamic environments would require motion-aware correspondence and Gaussian modeling.

\section{Conclusion}
We presented AdaGS, a confidence-gated diffusion adaptation framework for unposed sparse-view 3D Gaussian reconstruction. We identify an artifact amplification loop in which diffusion restoration converts Gaussian floaters into plausible but unsupported content that is subsequently absorbed back into the reconstruction. To interrupt this loop, AdaGS combines an artifact-aware adapter that suppresses misleading cues before restoration, bidirectional confidence fusion that gates generated content according to cross-view support, and representation-side Gaussian stabilization that attenuates persistent artifact sources. Experiments on DL3DV, Waymo, and KITTI demonstrate consistent improvements in novel-view synthesis and camera pose estimation over existing unposed reconstruction methods. Future work will extend the framework to dynamic scenes and more challenging low-overlap settings.

\bibliography{aaai2027}

@String(CVPR  = {IEEE Conf. Comput. Vis. Pattern Recog.})

@String(AAAI  = {AAAI})

@String(TOG   = {ACM Trans. Graph.})

@String(CVPR  = {CVPR})

@String(TOG   = {ACM TOG})

@article{kerbl20233dgs,
  title={3D Gaussian splatting for real-time radiance field rendering.},
  author={Kerbl, Bernhard and Kopanas, Georgios and Leimk{\"u}hler, Thomas and Drettakis, George},
  journal={ACM Trans. Graph.},
  volume={42},
  number={4},
  pages={139--1},
  year={2023}
}

@inproceedings{schonberger2016sfm,
  title={Structure-from-motion revisited},
  author={Schonberger, Johannes L and Frahm, Jan-Michael},
  booktitle={Proceedings of the IEEE conference on computer vision and pattern recognition},
  pages={4104--4113},
  year={2016}
}

@inproceedings{fu2024cf-3dgs,
  title={Colmap-free 3d gaussian splatting},
  author={Fu, Yang and Liu, Sifei and Kulkarni, Amey and Kautz, Jan and Efros, Alexei A and Wang, Xiaolong},
  booktitle={Proceedings of the IEEE/CVF Conference on Computer Vision and Pattern Recognition},
  pages={20796--20805},
  year={2024}
}

@inproceedings{jiang2024cogs,
  title={A construct-optimize approach to sparse view synthesis without camera pose},
  author={Jiang, Kaiwen and Fu, Yang and Varma T, Mukund and Belhe, Yash and Wang, Xiaolong and Su, Hao and Ramamoorthi, Ravi},
  booktitle={ACM SIGGRAPH 2024 Conference Papers},
  pages={1--11},
  year={2024}
}

@article{dong2025rob-gs,
  title={Towards Better Robustness: Progressively Joint Pose-3DGS Learning for Arbitrarily Long Videos},
  author={Dong, Zhen-Hui and Ye, Sheng and Wen, Yu-Hui and Li, Nannan and Liu, Yong-Jin},
  journal={arXiv e-prints},
  pages={arXiv--2501},
  year={2025}
}

@article{shi2025trackgs,
  title={Trackgs: Optimizing colmap-free 3d gaussian splatting with global track constraints},
  author={Shi, Dongbo and Cao, Shen and Fan, Lubin and Wu, Bojian and Guo, Jinhui and Chen, Renjie and Liu, Ligang and Ye, Jieping},
  journal={arXiv preprint arXiv:2502.19800},
  year={2025}
}

@inproceedings{matsuki2024monogs,
  title={Gaussian splatting slam},
  author={Matsuki, Hidenobu and Murai, Riku and Kelly, Paul HJ and Davison, Andrew J},
  booktitle={Proceedings of the IEEE/CVF Conference on Computer Vision and Pattern Recognition},
  pages={18039--18048},
  year={2024}
}

@article{yu2025opengs-slam,
  title={Rgb-only gaussian splatting slam for unbounded outdoor scenes},
  author={Yu, Sicheng and Cheng, Chong and Zhou, Yifan and Yang, Xiaojun and Wang, Hao},
  journal={arXiv preprint arXiv:2502.15633},
  year={2025}
}

@inproceedings{cheng2025s3po-gs,
  title={Outdoor monocular slam with global scale-consistent 3d gaussian pointmaps},
  author={Cheng, Chong and Yu, Sicheng and Wang, Zijian and Zhou, Yifan and Wang, Hao},
  booktitle={Proceedings of the IEEE/CVF International Conference on Computer Vision},
  pages={26035--26044},
  year={2025}
}

@inproceedings{sandstrom2025splat-slam,
  title={Splat-slam: Globally optimized rgb-only slam with 3d gaussians},
  author={Sandstr{\"o}m, Erik and Zhang, Ganlin and Tateno, Keisuke and Oechsle, Michael and Niemeyer, Michael and Zhang, Youmin and Patel, Manthan and Van Gool, Luc and Oswald, Martin and Tombari, Federico},
  booktitle={Proceedings of the Computer Vision and Pattern Recognition Conference},
  pages={1680--1691},
  year={2025}
}

@inproceedings{wang2024dust3r,
  title={Dust3r: Geometric 3d vision made easy},
  author={Wang, Shuzhe and Leroy, Vincent and Cabon, Yohann and Chidlovskii, Boris and Revaud, Jerome},
  booktitle={Proceedings of the IEEE/CVF Conference on Computer Vision and Pattern Recognition},
  pages={20697--20709},
  year={2024}
}

@inproceedings{leroy2024mast3r,
  title={Grounding image matching in 3d with mast3r},
  author={Leroy, Vincent and Cabon, Yohann and Revaud, J{\'e}r{\^o}me},
  booktitle={European Conference on Computer Vision},
  pages={71--91},
  year={2024},
  organization={Springer}
}

@inproceedings{wang2025vggt,
  title={Vggt: Visual geometry grounded transformer},
  author={Wang, Jianyuan and Chen, Minghao and Karaev, Nikita and Vedaldi, Andrea and Rupprecht, Christian and Novotny, David},
  booktitle={Proceedings of the Computer Vision and Pattern Recognition Conference},
  pages={5294--5306},
  year={2025}
}

@article{fan2024instantsplat,
  title={Instantsplat: Sparse-view gaussian splatting in seconds},
  author={Fan, Zhiwen and Cong, Wenyan and Wen, Kairun and Wang, Kevin and Zhang, Jian and Ding, Xinghao and Xu, Danfei and Ivanovic, Boris and Pavone, Marco and Pavlakos, Georgios and others},
  journal={arXiv preprint arXiv:2403.20309},
  year={2024}
}

@inproceedings{lin2025longsplat,
  title={Longsplat: Robust unposed 3d gaussian splatting for casual long videos},
  author={Lin, Chin-Yang and Sun, Cheng and Yang, Fu-En and Chen, Min-Hung and Lin, Yen-Yu and Liu, Yu-Lun},
  booktitle={Proceedings of the IEEE/CVF International Conference on Computer Vision},
  pages={27412--27422},
  year={2025}
}

@article{cheng2025unposed,
  title={Unposed 3DGS Reconstruction with Probabilistic Procrustes Mapping},
  author={Cheng, Chong and Wang, Zijian and Yu, Sicheng and Hu, Yu and Yao, Nanjie and Wang, Hao},
  journal={arXiv preprint arXiv:2507.18541},
  year={2025}
}

@inproceedings{li2024dngaussian,
  title={Dngaussian: Optimizing sparse-view 3d gaussian radiance fields with global-local depth normalization},
  author={Li, Jiahe and Zhang, Jiawei and Bai, Xiao and Zheng, Jin and Ning, Xin and Zhou, Jun and Gu, Lin},
  booktitle={Proceedings of the IEEE/CVF conference on computer vision and pattern recognition},
  pages={20775--20785},
  year={2024}
}

@inproceedings{xu2025depthsplat,
  title={Depthsplat: Connecting gaussian splatting and depth},
  author={Xu, Haofei and Peng, Songyou and Wang, Fangjinhua and Blum, Hermann and Barath, Daniel and Geiger, Andreas and Pollefeys, Marc},
  booktitle={Proceedings of the Computer Vision and Pattern Recognition Conference},
  pages={16453--16463},
  year={2025}
}

@inproceedings{zhang2025transplat,
  title={Transplat: Generalizable 3d gaussian splatting from sparse multi-view images with transformers},
  author={Zhang, Chuanrui and Zou, Yingshuang and Li, Zhuoling and Yi, Minmin and Wang, Haoqian},
  booktitle={Proceedings of the AAAI Conference on Artificial Intelligence},
  volume={39},
  number={9},
  pages={9869--9877},
  year={2025}
}

@inproceedings{paliwal2024coherentgs,
  title={Coherentgs: Sparse novel view synthesis with coherent 3d gaussians},
  author={Paliwal, Avinash and Ye, Wei and Xiong, Jinhui and Kotovenko, Dmytro and Ranjan, Rakesh and Chandra, Vikas and Kalantari, Nima Khademi},
  booktitle={European Conference on Computer Vision},
  pages={19--37},
  year={2024},
  organization={Springer}
}

@inproceedings{zheng2025nexusgs,
  title={NexusGS: Sparse View Synthesis with Epipolar Depth Priors in 3D Gaussian Splatting},
  author={Zheng, Yulong and Jiang, Zicheng and He, Shengfeng and Sun, Yandu and Dong, Junyu and Zhang, Huaidong and Du, Yong},
  booktitle={Proceedings of the Computer Vision and Pattern Recognition Conference},
  pages={26800--26809},
  year={2025}
}

@inproceedings{chung2024depth-regular,
  title={Depth-regularized optimization for 3d gaussian splatting in few-shot images},
  author={Chung, Jaeyoung and Oh, Jeongtaek and Lee, Kyoung Mu},
  booktitle={Proceedings of the IEEE/CVF Conference on Computer Vision and Pattern Recognition},
  pages={811--820},
  year={2024}
}

@article{liu20243dgs-enhancer,
  title={3dgs-enhancer: Enhancing unbounded 3d gaussian splatting with view-consistent 2d diffusion priors},
  author={Liu, Xi and Zhou, Chaoyi and Huang, Siyu},
  journal={Advances in Neural Information Processing Systems},
  volume={37},
  pages={133305--133327},
  year={2024}
}

@inproceedings{wu2025difix3d+,
  title={Difix3d+: Improving 3d reconstructions with single-step diffusion models},
  author={Wu, Jay Zhangjie and Zhang, Yuxuan and Turki, Haithem and Ren, Xuanchi and Gao, Jun and Shou, Mike Zheng and Fidler, Sanja and Gojcic, Zan and Ling, Huan},
  booktitle={Proceedings of the Computer Vision and Pattern Recognition Conference},
  pages={26024--26035},
  year={2025}
}

@inproceedings{zhong2025taming-video,
  title={Taming Video Diffusion Prior with Scene-Grounding Guidance for 3D Gaussian Splatting from Sparse Inputs},
  author={Zhong, Yingji and Li, Zhihao and Chen, Dave Zhenyu and Hong, Lanqing and Xu, Dan},
  booktitle={Proceedings of the Computer Vision and Pattern Recognition Conference},
  pages={6133--6143},
  year={2025}
}

@article{yu2024viewcrafter,
  title={Viewcrafter: Taming video diffusion models for high-fidelity novel view synthesis},
  author={Yu, Wangbo and Xing, Jinbo and Yuan, Li and Hu, Wenbo and Li, Xiaoyu and Huang, Zhipeng and Gao, Xiangjun and Wong, Tien-Tsin and Shan, Ying and Tian, Yonghong},
  journal={arXiv preprint arXiv:2409.02048},
  year={2024}
}

@article{liu2024reconx,
  title={Reconx: Reconstruct any scene from sparse views with video diffusion model},
  author={Liu, Fangfu and Sun, Wenqiang and Wang, Hanyang and Wang, Yikai and Sun, Haowen and Ye, Junliang and Zhang, Jun and Duan, Yueqi},
  journal={arXiv preprint arXiv:2408.16767},
  year={2024}
}

@article{han2024binocular,
  title={Binocular-guided 3d gaussian splatting with view consistency for sparse view synthesis},
  author={Han, Liang and Zhou, Junsheng and Liu, Yu-Shen and Han, Zhizhong},
  journal={Advances in Neural Information Processing Systems},
  volume={37},
  pages={68595--68621},
  year={2024}
}

@inproceedings{zhang2024cor-gs,
  title={Cor-gs: sparse-view 3d gaussian splatting via co-regularization},
  author={Zhang, Jiawei and Li, Jiahe and Yu, Xiaohan and Huang, Lei and Gu, Lin and Zheng, Jin and Bai, Xiao},
  booktitle={European Conference on Computer Vision},
  pages={335--352},
  year={2024},
  organization={Springer}
}

@inproceedings{xu2025dropoutgs,
  title={DropoutGS: Dropping Out Gaussians for Better Sparse-view Rendering},
  author={Xu, Yexing and Wang, Longguang and Chen, Minglin and Ao, Sheng and Li, Li and Guo, Yulan},
  booktitle={Proceedings of the Computer Vision and Pattern Recognition Conference},
  pages={701--710},
  year={2025}
}

@inproceedings{park2025dropgaussian,
  title={Dropgaussian: Structural regularization for sparse-view gaussian splatting},
  author={Park, Hyunwoo and Ryu, Gun and Kim, Wonjun},
  booktitle={Proceedings of the Computer Vision and Pattern Recognition Conference},
  pages={21600--21609},
  year={2025}
}

@article{chen2025quantifying,
  title={Quantifying and Alleviating Co-Adaptation in Sparse-View 3D Gaussian Splatting},
  author={Chen, Kangjie and Zhong, Yingji and Li, Zhihao and Lin, Jiaqi and Chen, Youyu and Qin, Minghan and Wang, Haoqian},
  journal={arXiv preprint arXiv:2508.12720},
  year={2025}
}

@inproceedings{cheng2025reggs,
  title={RegGS: Unposed Sparse Views Gaussian Splatting with 3DGS Registration},
  author={Cheng, Chong and Hu, Yu and Yu, Sicheng and Zhao, Beizhen and Wang, Zijian and Wang, Hao},
  booktitle={Proceedings of the IEEE/CVF International Conference on Computer Vision},
  pages={8100--8109},
  year={2025}
}

@article{ye2024noposplat,
  title={No pose, no problem: Surprisingly simple 3d gaussian splats from sparse unposed images},
  author={Ye, Botao and Liu, Sifei and Xu, Haofei and Li, Xueting and Pollefeys, Marc and Yang, Ming-Hsuan and Peng, Songyou},
  journal={arXiv preprint arXiv:2410.24207},
  year={2024}
}

@inproceedings{yu2025sgd,
  title={Sgd: Street view synthesis with gaussian splatting and diffusion prior},
  author={Yu, Zhongrui and Wang, Haoran and Yang, Jinze and Wang, Hanzhang and Cao, Jiale and Ji, Zhong and Sun, Mingming},
  booktitle={2025 IEEE/CVF Winter Conference on Applications of Computer Vision (WACV)},
  pages={3812--3822},
  year={2025},
  organization={IEEE}
}

@article{paul2024gaussianscene,
  title={Gaussian Scenes: Pose-Free Sparse-View Scene Reconstruction using Depth-Enhanced Diffusion Priors},
  author={Paul, Soumava and Kaushik, Prakhar and Yuille, Alan},
  journal={arXiv preprint arXiv:2411.15966},
  year={2024}
}

@inproceedings{lu2024scaffold-gs,
  title={Scaffold-gs: Structured 3d gaussians for view-adaptive rendering},
  author={Lu, Tao and Yu, Mulin and Xu, Linning and Xiangli, Yuanbo and Wang, Limin and Lin, Dahua and Dai, Bo},
  booktitle={Proceedings of the IEEE/CVF Conference on Computer Vision and Pattern Recognition},
  pages={20654--20664},
  year={2024}
}

@inproceedings{zhao2025wavelet,
  title={Wavelet-GS: 3D Gaussian Splatting with Wavelet Decomposition},
  author={Zhao, Beizhen and Zhou, Yifan and Yu, Sicheng and Wang, Zijian and Wang, Hao},
  booktitle={Proceedings of the 33rd ACM International Conference on Multimedia},
  pages={8616--8625},
  year={2025}
}

@InProceedings{Sun_2020_CVPRwaymo, author = {Sun, Pei and Kretzschmar, Henrik and Dotiwalla, Xerxes and Chouard, Aurelien and Patnaik, Vijaysai and Tsui, Paul and Guo, James and Zhou, Yin and Chai, Yuning and Caine, Benjamin and Vasudevan, Vijay and Han, Wei and Ngiam, Jiquan and Zhao, Hang and Timofeev, Aleksei and Ettinger, Scott and Krivokon, Maxim and Gao, Amy and Joshi, Aditya and Zhang, Yu and Shlens, Jonathon and Chen, Zhifeng and Anguelov, Dragomir}, title = {Scalability in Perception for Autonomous Driving: Waymo Open Dataset}, booktitle = {Proceedings of the IEEE/CVF Conference on Computer Vision and Pattern Recognition (CVPR)}, month = {June}, year = {2020} }

@article{geiger2013kittidata,
  title={Vision meets robotics: The kitti dataset},
  author={Geiger, Andreas and Lenz, Philip and Stiller, Christoph and Urtasun, Raquel},
  journal={The international journal of robotics research},
  volume={32},
  number={11},
  pages={1231--1237},
  year={2013},
  publisher={Sage Publications Sage UK: London, England}
}

@inproceedings{ling2024dl3dv,
  title={Dl3dv-10k: A large-scale scene dataset for deep learning-based 3d vision},
  author={Ling, Lu and Sheng, Yichen and Tu, Zhi and Zhao, Wentian and Xin, Cheng and Wan, Kun and Yu, Lantao and Guo, Qianyu and Yu, Zixun and Lu, Yawen and others},
  booktitle={Proceedings of the IEEE/CVF Conference on Computer Vision and Pattern Recognition},
  pages={22160--22169},
  year={2024}
}

@article{cong2025videolifter,
  title={Videolifter: Lifting videos to 3d with fast hierarchical stereo alignment},
  author={Cong, Wenyan and Zhu, Hanqing and Wang, Kevin and Lei, Jiahui and Stearns, Colton and Cai, Yuanhao and Wang, Dilin and Ranjan, Rakesh and Feiszli, Matt and Guibas, Leonidas and others},
  journal={arXiv preprint arXiv:2501.01949},
  year={2025}
}

@article{chen2025dggt,
  title={DGGT: Feedforward 4D Reconstruction of Dynamic Driving Scenes using Unposed Images},
  author={Chen, Xiaoxue and Xiong, Ziyi and Chen, Yuantao and Li, Gen and Wang, Nan and Luo, Hongcheng and Chen, Long and Sun, Haiyang and Wang, Bing and Chen, Guang and others},
  journal={arXiv preprint arXiv:2512.03004},
  year={2025}
}

@article{jiang2025anysplat,
  title={Anysplat: Feed-forward 3d gaussian splatting from unconstrained views},
  author={Jiang, Lihan and Mao, Yucheng and Xu, Linning and Lu, Tao and Ren, Kerui and Jin, Yichen and Xu, Xudong and Yu, Mulin and Pang, Jiangmiao and Zhao, Feng and others},
  journal={ACM Transactions on Graphics (TOG)},
  volume={44},
  number={6},
  pages={1--16},
  year={2025},
  publisher={ACM New York, NY, USA}
}

@article{lin2025depth,
  title={Depth anything 3: Recovering the visual space from any views},
  author={Lin, Haotong and Chen, Sili and Liew, Junhao and Chen, Donny Y and Li, Zhenyu and Shi, Guang and Feng, Jiashi and Kang, Bingyi},
  journal={arXiv preprint arXiv:2511.10647},
  year={2025}
}

@article{lu2025prosplat,
  title={ProSplat: Improved Feed-Forward 3D Gaussian Splatting for Wide-Baseline Sparse Views},
  author={Lu, Xiaohan and Fu, Jiaye and Zhang, Jiaqi and Song, Zetian and Jia, Chuanmin and Ma, Siwei},
  journal={arXiv preprint arXiv:2506.07670},
  year={2025}
}

\end{document}